\documentclass[letterpaper, 10 pt, conference]{ieeeconf}  
\usepackage{amsmath,amsfonts}
\usepackage{algorithmic}
\usepackage{algorithm}
\usepackage{array}
\usepackage[caption=false,font=normalsize,labelfont=sf,textfont=sf]{subfig}
\usepackage{textcomp}
\usepackage{stfloats}
\usepackage{url}
\usepackage{verbatim}
\usepackage{graphicx}
\usepackage{cite}
\usepackage{paralist}
\usepackage{multirow}
\usepackage{tabularx}
\usepackage[hidelinks]{hyperref}
\usepackage[table]{xcolor}

\usepackage{color, colortbl}
\definecolor{LightCyan}{rgb}{0.88,1,1}

\IEEEoverridecommandlockouts                              

\title{\bf
Directionality-Aware Mixture Model Parallel Sampling for \\ Efficient Linear Parameter Varying Dynamical System Learning}

\author{Sunan Sun$^*$, Haihui Gao, Tianyu Li and Nadia Figueroa
\thanks{All authors are with the Department of Mechanical Engineering, University of Pennsylvania,
Philadelphia PA 19104, USA}
\thanks{$^*$Corresponding author. (e-mail: sunan@seas.upenn.edu)}
}

\begin{document}
\maketitle
\thispagestyle{empty}
\pagestyle{empty}

\begin{abstract}
The Linear Parameter Varying Dynamical System (LPV-DS) is an effective approach that learns stable, time-invariant motion policies  using statistical modeling and semi-definite optimization to encode complex motions for reactive robot control. Despite its strengths, the LPV-DS learning approach faces challenges in achieving a high model accuracy without compromising the computational efficiency. To address this, we introduce the Directionality-Aware Mixture Model (DAMM), a novel statistical model that applies the Riemannian metric on the n-sphere $\mathbb{S}^n$ to efficiently blend non-Euclidean directional data with $\mathbb{R}^m$ Euclidean states. Additionally, we develop a hybrid Markov chain Monte Carlo technique that combines Gibbs Sampling with Split/Merge Proposal, allowing for parallel computation to drastically speed up inference. Our extensive empirical tests demonstrate that LPV-DS integrated with DAMM achieves higher reproduction accuracy, better model efficiency, and near real-time/online learning compared to standard estimation methods on various datasets. Lastly, we demonstrate its suitability for incrementally learning multi-behavior policies in real-world robot experiments.
\end{abstract}

\section{Introduction} \label{sec:intro}
Safe integration of robots into human workspaces requires the ability to adapt and replan in response to changing environments and constraints. Traditional path planning approaches, assuming a known environment and robot dynamics, face challenges when confronted with uncertainties and perturbations during operation~\cite{UDE1993113, Mobilerobotsat, ALEOTTI2006409}. In contrast, Dynamical System (DS)-based motion policies leverage redundancy of solutions in dynamic environments, embedding an infinite set of feasible solutions in a single control law to overcome environmental uncertainties and perturbations~\cite{TEXTBOOK}. Furthermore, stability conditions can be introduced as constraints in the learning of DS, providing a closed-form analytical solution to trajectory planning~\cite{gribovskaya2011learning, SEDS}.

Our focus is on learning stable, time-independent motion policies from limited demonstrations, emphasizing i) state-space coverage, ii) minimal training data, iii) model accuracy, and iv) computational efficiency for online and incremental learning.  While recent neural network (NN) based formulations for stable DS motion policies show promising results in encoding highly non-linear trajectories; as adopting normalizing flows~\cite{9341035}, euclideanizing flows~\cite{pmlr-v120-rana20a} or via contrastive learning~\cite{10214439}; such NN-based methods require many trajectories and substantial computation time to reach stable solutions. Interestingly, the seminal works on the Linear Parameter Varying Dynamical System (LPV-DS) formulation and its Gaussian Mixture Model (GMM) based learning frameworks~\cite{SEDS, PC-GMM, TEXTBOOK} can achieve objectives i-ii), but face the challenge of preserving high reproduction accuracy without compromising computational speed.

\begin{figure}[!t]
\centering
\includegraphics[width=0.88\linewidth]{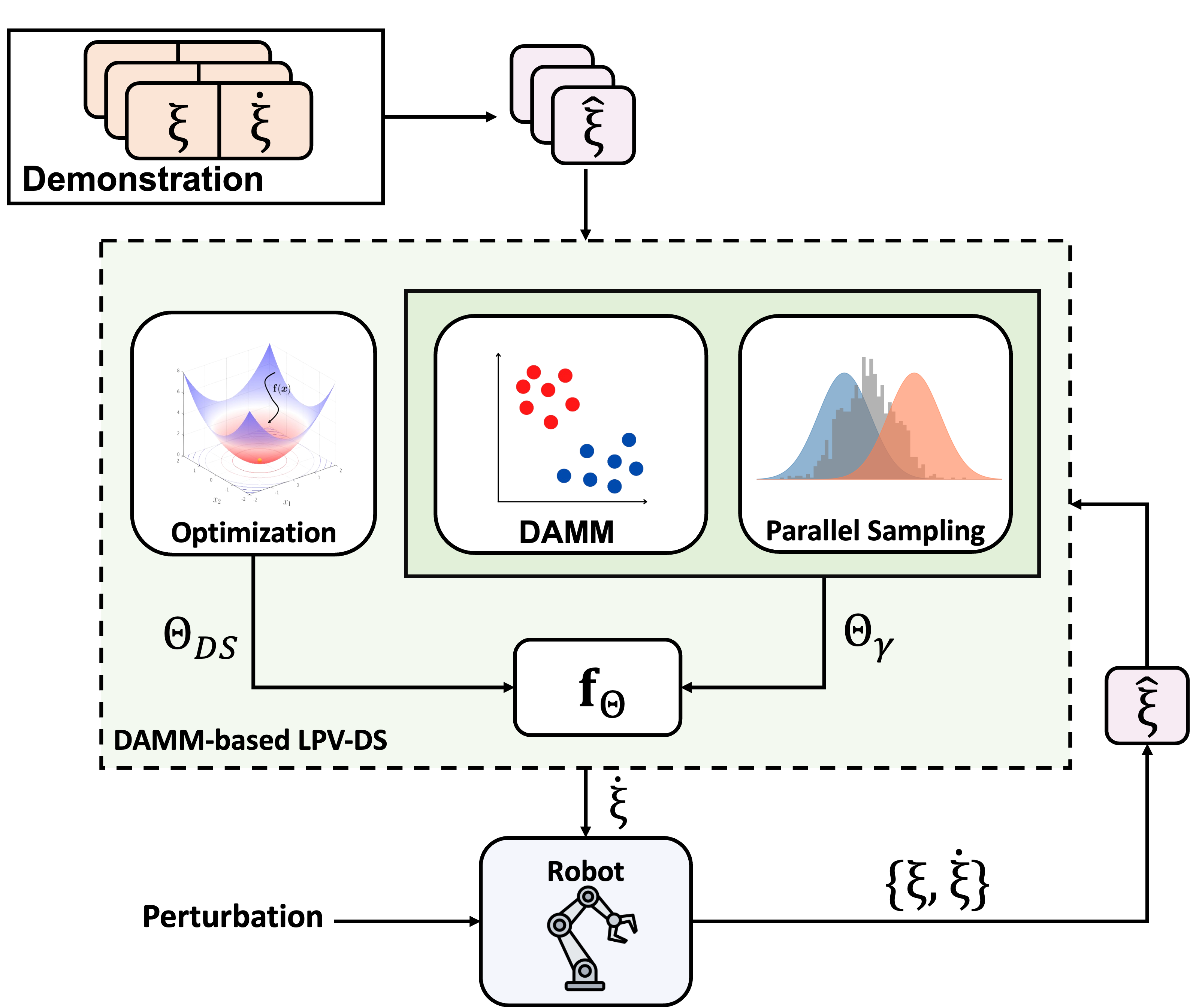}
\caption{The schematic of the \textbf{DAMM-based LPV-DS} formulation which consists of \textbf{DAMM} and \textbf{Parallel Sampling} to cluster and parameterize the trajectory, and the optimization to minimize the prediction error; the resulting DS, $f_\Theta$, takes the position $\xi$ and velocity $\Dot{\xi}$ as inputs, transforms them into the augmented state $\Hat{\xi}$, and generates the estimated desired linear velocity which is then passed down to command the robot via a low-level feedback controller; e.g. a Cartesian twist impedance controller.}\label{fig:pipeline}
\vspace{-15pt}
\end{figure}

While GMM is useful at clustering sparse points, it fails to produce physically-meaningful representation of trajectory data. In the applications of trajectory planning/control, trajectories have inherent directionality; however, GMMs do not explicitly model directionality. Despite its fast inference via standard Gibbs sampling, a GMM struggles to encapsulate the intrinsic motion and dynamic nature of trajectory data, resulting in erroneous DS. To alleviate these issues, Physically Consistent (PC)-GMM was proposed \cite{PC-GMM}; a state-of-the-art statistical model tailored to the LPV-DS framework. By applying a distance-dependent Chinese Restaurant Process (DD-CRP) prior~\cite{DD-CRP}, PC-GMM integrates a distance metric of directionality by computing the pair-wise cosine similarity between every observation. Considering the directionality as side-information, PC-GMM produces more informative clustering results and DS. However, PC-GMM suffers from slow inference due to the DD-CRP requiring the computation of the similarity matrix, resulting in memory inefficiency and exponential increase in computation time wrt. the data size. Moreover, the online learning of PC-GMM is hard to achieve because the inference with the DD-CRP prior necessitates incremental updates~\cite{DD-CRP}, ruling out the possibility of parallel computation as such updates are \textit{strictly sequential}.

In tackling the aforementioned challenges, we introduce the \textbf{Directionality-Aware Mixture Model (DAMM)}. Inspired by relevant work on clustering spherical data on Riemannian manifolds \cite{SPHERE}, the DAMM formulation incorporates directional information using a proper Riemannian metric on the directional data manifold, inherently capturing the directionality within trajectories and producing physically-meaningful DS (Section \ref{sec:damm}). We then introduce a new parallel Markov chain Monte Carlo (MCMC) sampling scheme tailored to the DAMM formulation, that is capable of achieving online performance (Section \ref{sec:sampling}). 

We evaluate our approach through extensive empirical validation, including benchmark comparisons on LASA datasets \cite{khansari2014learning} and the PC-GMM dataset \cite{PC-GMM} (including 2D and 3D real trajectories) against PC-GMM and baseline methods (Section \ref{sec:result}). We demonstrate that the DAMM-based LPV-DS framework exhibits enhanced capabilities in producing improved DS across various metrics, and faster learning speed than its predecessors by order of magnitude. We further validate our approach in the real robot experiments (Section \ref{sec:robot}), where a single batch trajectory of 500 observations can be learned in $<500ms$, making our approach, to the best of our knowledge, the first DS learning framework that can be estimated in near real-time scale. The schematic of DAMM-based LPV-DS in a typical robotic control workflow is shown in Fig.~\ref{fig:pipeline}. In the next section, we revisit the LPV-DS formulation, and give a brief overview of unit sphere geometry and Gibbs sampling.

\section{Preliminaries}\label{sec:prelims}
\subsection{The LPV-DS Formulation\label{sec:lpv-ds} }
Let $\xi, \dot{\xi} \in \mathbb{R}^d$ represent the kinematic robot state and velocity vectors. In the DS-based motion policy literature \cite{TEXTBOOK}, $\dot{\xi} = f(\xi)$ is a first-order DS that describes a motion policy in the robot's state space $\mathbb{R}^d$. The goal of DS-based learning from demonstration (LfD) is to infer $f(\xi): \mathbb{R}^d \rightarrow \mathbb{R}^d$ from data, such that any point $\xi$ in the state space leads to a stable attractor $\xi^*\in \mathbb{R}^d$, with $f(\xi)$ described by a set of parameters $\Theta$ and attractor $\xi^*\in\mathbb{R}^d$; mathematically $\dot{\xi} = f(\xi; \Theta, \xi^*) \Rightarrow \lim_{t \to \infty} \|\xi - \xi^*\| = 0$, i.e., the DS is globally asymptotically stable (GAS)~\cite{Khalil:1173048}.

Learning $\Dot{\xi}=f(\xi)$ can be framed as a regression problem, where the inputs are the state variables $\xi$ and the outputs are the first-order time derivative $\Dot{\xi}$. Such formulation gives rise to the utilization of statistical methods for estimating the parameters $\Theta$. However, standard regression techniques cannot ensure globally asymptotic stability. To alleviate this, the LPV-DS approach was first introduced in the seminal work of \cite{SEDS} as a constrained Gaussian Mixture Regression (GMR) and then formalized as the untied GMM-based LPV-DS approach in \cite{PC-GMM}, where a nonlinear DS is encoded as a mixture of continuous linear time-invariant (LTI) systems: 
\begin{equation} \label{eq:lpv_ds}
\begin{aligned}
    &\quad \quad \dot{\xi}=f(\xi; \Theta)=\sum_{k=1}^K \gamma_k(\xi)\left(\bold{A}_k \xi+b_k\right)\\
    \text{s.t.}&\,\,  \left\{\begin{array}{l}
    \left(\bold{A}_k\right)^T \bold{P}+\bold{P} \bold{A}_k=\bold{Q}_k, \bold{Q}_k=\left(\bold{Q}_k\right)^T \prec 0 \\
    b_k=-\bold{A}_k \xi^* \end{array}\right.
\end{aligned}
\end{equation}
where $\gamma_k(\xi)$ is the state-dependent mixing function that quantifies the weight of each LTI system $(\bold{A}_k \xi + b_k)$ and $\Theta = \{\theta_\gamma\}_{\gamma=1}^K = \{\gamma_k, \bold{A}_k, b_k\}_{k=1}^K$ is the set of parameters to learn. The constraints of the Eq. \ref{eq:lpv_ds} enforce GAS of the result DS derived from a parametrized Lyapunov function $V(\xi) = (\xi-\xi^*)^T\bold{P}(\xi-\xi^*)$ with $\bold{P}=\bold{P}^T\succ0$~\cite{PC-GMM, TEXTBOOK}.

To ensure GAS of Eq. \ref{eq:lpv_ds}, besides enforcing the Lyapunov stability constraints on the LTI parameters one must ensure that $0<\gamma_k(\xi)< 1$ and $\sum_{k=1}^K \gamma_k(\xi)= 1 ~\forall \xi \in \mathbb{R}^d$. As noted in \cite{PC-GMM}, this is achieved by formulating $ \gamma_k({\xi}) = \frac{\pi_k \mathcal{N}({\xi}| \theta_k)}{\sum_{j=1}\pi_j \mathcal{N}(\xi| \theta_j)}$ as the \textit{a posteriori probability} of the state $\xi$ from a GMM used to partition the nonlinear DS into linear components. Here, $K$ is the number of components corresponding to the number of LTIs, $\mathcal{N}({\xi}| \theta_k)$ is the probability of observing ${\xi}$ from the $k$-th Gaussian component parametrized by mean and covariance matrix $\theta_k =\{\mu_k,\bold{\Sigma}_k \}$, and $\pi_k$ is the prior probability of an observation from this particular component satisfying $\sum_{k=1}^K \pi_k = 1$.

In \cite{PC-GMM} a two-step estimation framework was proposed to estimate the GMM parameters $\Theta_{\gamma}=\{\pi_k, \mu_k, \bold{\Sigma}_k\}_{k=1}^K$ and the DS parameters $\Theta_{DS} = \{\bold{A}_k, b_k\}_{k=1}^K$ forming $\Theta=\{\Theta_{\gamma}, \Theta_{DS}\}$. First, given the set of reference trajectories $\mathcal{D}:=\{\xi^{\mathrm{ref}}_i\ \dot{\xi}^{\mathrm{ref}}_i\}_{i = 1}^N$, where $i$ is the sequence order of the sampled states, a GMM is fit to the position variables of the reference trajectory, $\{\xi^{ref}_i\ \}_{i = 1}^N$,  to obtain $\Theta_{\gamma}$. The optimal number of Gaussians $K$ and their placement can be estimated by model selection via Expectation-Maximization or via Bayesian non-parametric estimation. Then, $\Theta_{DS}$ are learned through a semi-definite program minimizing reproduction accuracy subject to stability constraints \cite{PC-GMM,TEXTBOOK}.
\vspace{-3pt}
\subsection{$n$-Sphere Geometry Overview}
\vspace{-3pt}
A $n$-dimensional hypersphere with a radius of 1, known as the $n$-sphere or $\mathbb{S}^n$, is a Riemannian manifold embedded in $n+1$-dimensional Euclidean space $\mathbb{R}^{n+1}$. A Riemannian manifold is a smooth manifold equipped with positive definite inner product defined in the tangent space at each point. This metric allows for the measurement of distances, angles, and other geometric properties on the manifold. For clarity, we denote elements of the manifold in bold and elements in tangent space in fraktur typeface; i.e. $\bold{q} \in \mathcal{M}$ and $\frak{q} \in T_\bold{p}\mathcal{M}$. 

\begin{figure}[!t]
\centering
\includegraphics[width=1\linewidth]{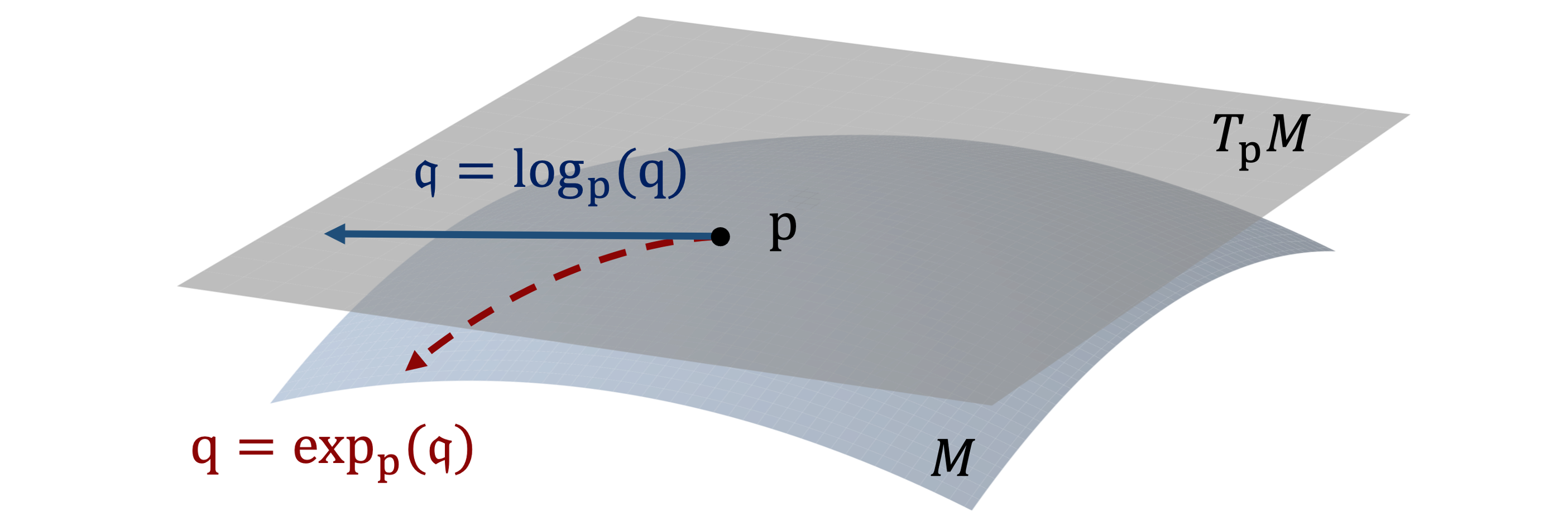}%
\hfill
\caption{Illustrative example of the exponential/logarithmic mapping on a Riemannian manifold and its tangent space defined at point $p$}
\label{fig:riem}
\vspace{-10pt}
\end{figure}

The notion of distance on unit sphere is a generalization of straight lines in Euclidean spaces. The minimum distance paths that lie on the curve, also called geodesics, are defined as $d(\bold{p}, \bold{q}) = \arccos(\bold{p}^T\bold{q})$ between two points on unit sphere, or $\bold{p}, \bold{q}\in \mathbb{S}^d$~\cite{RIEM-GEO, RIEM-GEO2}. We can also compute the Riemannian equivalent of mean and covariance as follows,
\begin{equation} \label{eq:riem_mu_sigma}
\begin{aligned}
            &\quad \Tilde{\mu} = \operatornamewithlimits{argmin}_{\bold{p} \in \mathbb{S}^{d}} \sum_{i=1} ^N  d(\bold{q}_i,\ \bold{p})^2\\
            \Tilde{\bold{\Sigma}}&= \frac{1}{(N-1)} \sum_{i=1}^N \log_{\Tilde{\mu}}(\bold{p}_i)\log_{\Tilde{\mu}}(\bold{p}_i)^T.
\end{aligned}
\end{equation}
The average $\Tilde{\mu}$, defined as the center of mass on unit sphere, employs the notion of the Fréchet mean~\cite{RIEM-MEAN}, which extends the sample mean from $\mathbb{R}^d$ to Riemannian manifolds $\mathcal{M}$. In practice, $\Tilde{\mu}$ can be efficiently computed in an iterative approach~\cite{RIEM-COV}. The empirical covariance $\Tilde{\bold{\Sigma}}$ captures the dispersion of data in tangent space $T_\bold{p}\mathcal{M}$, where the logarithmic map $\log_\bold{p}:\mathcal{M} \rightarrow  T_\bold{p}\mathcal{M}$ maps a point on the Riemannian manifold to the tangent space defined by the point of tangency $\bold{p}$:
\begin{equation} \label{eq:riem_log}
    \frak{q} = \log_\bold{p}(\bold{q}) = d(\bold{p}, \bold{q}) \frac{\bold{q} - \bold{p}^T\bold{q}\bold{p}}{\|\bold{q} - \bold{p}^T\bold{q}\bold{p}\|}.
\end{equation}
The inverse map is the exponential map $\exp_\bold{p}: T_\bold{p}\mathcal{M} \rightarrow \mathcal{M}$ which maps a point in tangent space of $\bold{p}$ to the manifold so that the mapped point lies in the direction of the geodesic starting at $\bold{p}$~\cite{RIEM-COV, RIEM-STAT, RIEM-STAT-2}:
\begin{equation} \label{eq:riem_exp}
    \bold{q} = \exp_\bold{p}(\frak{q}) = \bold{p}\cos(\| \frak{q} \|) + \frac{\frak{q}}{\| \frak{q} \|}\sin(\| \frak{q} \|).
\end{equation}
The L2 norm $\| \log_{\bold{p}} (\bold{q}) \|_2$ in the tangent space $T\mathbb{S}^d$ is equal to the geodesics between $\bold{p}$ and $\bold{q}$ on the manifold: $d(\bold{p}, \bold{q})$. However, this is true only when $\bold{p}$ is the point of tangency. In general, the distance between two other points in $T\mathbb{S}^d$ is not equal to the geodesic distance between their corresponding points in $\mathbb{S}^d$. An illustration of Riemannian operation on the manifold is shown in Fig.~\ref{fig:riem}.

\subsection{Gibbs Sampler Overview\label{sec:gibbs}}
Gibbs sampling is a Markov Chain Monte Carlo (MCMC) technique used for inference in probabilistic models. By iteratively sampling from posterior distribution, Gibbs sampler can estimate unknown parameters given observed data. Gibbs samplers fall under two categories: i) collapsed-weight (CW) Gibbs sampler and ii) instantiated-weight (IW) Gibbs sampler~\cite{GIBBS}. The CW Gibbs sampler (used in PC-GMM \cite{PC-GMM}) marginalizes out parameters and incrementally update each data point~\cite{DD-CRP}. On the other hand, IW Gibbs sampler instantiates all parameters in the beginning of every iteration and draws samples at once. We demonstrate IW Gibbs sampler using the inference of GMM as an example: 
\begin{equation} \label{eq:iw_gibbs}
        \begin{gathered}
            (\pi_1, \dots, \pi_K) \sim Cat(N_1, \dots, N_K), \\
            (\mu_k, \bold{\Sigma}_k) \sim  NIW(\Psi_n, \nu_n, \mu_n, \kappa_n), \ \forall k\in \{1,\dots, K\}\\
            z_i \overset{\propto}{\sim} \sum_{k=1}^K \pi_k \mathcal{N}(\xi_i \vert \mu_k, \bold{\Sigma}_k), \quad \forall i\in \{1,\dots, N\},
        \end{gathered}    
\end{equation}
in the beginning of each iteration, IW Gibbs sampler draws the cluster proportion $\pi_k$ from a categorical distribution defined by the number of observations $N_k$ in the $k$-th component, then samples the parameters of each Gaussian from the conjugate prior Normal-Inverse-Wishart (NIW) distribution defined by the posterior hyperparameters $(\Psi_n, \nu_n, \mu_n, \kappa_n)$~\cite{CONJUGATE}, and lastly draws the hidden variable or the assignment $z_i$ of each observation proportional to the cluster proportion $\pi$ and the posterior probability wrt. each component $N(\cdot \vert \theta)$. Note that each step in Eq.~\ref{eq:iw_gibbs} can be sampled in parallel across each component and each data; however, IW Gibbs sampler cannot create new components due to the finite-length instantiation.

\section{Directionality-Aware Mixture Model Parallel Sampling}\label{sec:method} 

\subsection{Directionality-Aware Mixture Model (DAMM)}
\label{sec:damm}
The DAMM formulation incorporates the directionality of trajectory data by identifying and segmenting non-linear trajectories into piece-wise linear components. Given the demonstration trajectory, we begin by normalizing the velocity vector $\Dot{\xi}$ and obtaining a unit-norm directional vector for each observation in $\mathcal{D}$ as $\xi^{dir} = \frac{\Dot{\xi}}{||\Dot{\xi} ||} \in \mathbb{S}^{d-1} \subset \mathbb{R}^d$. We note that $\xi^{dir}$, which represents the instantaneous direction, lies on the $(d-1)$-dimensional unit sphere or $\mathbb{S}^{d-1}$. Rather than computing the empirical covariance $\Tilde{\bold{\Sigma}}$ defined in Eq.~\ref{eq:riem_mu_sigma}, we construct a scalar-valued variance as follows:
\begin{equation} \label{eq:sigma_dir}
    (\sigma^2)^{dir} = \frac{1}{N-1} \sum_{i=1}^N || \log_{\mu^{dir}}(\xi^{dir}_i) ||_2^2 ,
\end{equation}
where $N$ is the number of data, $\mu^{dir}$ is the directional mean defined by Eq.~\ref{eq:riem_mu_sigma} and the Logarithmic mapping is defined by Eq.~\ref{eq:riem_log}. As opposed to $\Tilde{\bold{\Sigma}}$ which fully captures the variation with respect to \textit{all} directions in the manifold's geometry, $(\sigma^2)^{dir}$ describes the variation of direction relative to mean \textit{in terms of magnitude}. $(\sigma^2)^{dir}$ is favored over its higher dimensional counterpart because we are not interested in \textit{how} directions vary in the manifold, but how \textit{much} they vary within a component.  In other words, if a component contains a large variance $(\sigma^2)^{dir}$, then the trajectory associated with this component varies greatly in direction, hence considered non-linear, and should be split in a way that the new clusters retain lower variances $(\sigma^2)^{dir}$, resembling more linear components.

This gives rise to the new formulation of Gaussian component with its probability density and state defined as follow:
\begin{equation} \label{eq:DAMM}
        \mathcal{N}\bigg( \ \Hat{\xi}\ \bigg| \ \Hat{\mu} = 
        \begin{bmatrix}
            \mu^{pos}\\
            0
        \end{bmatrix},
        \Hat{\bold{\Sigma}} =  \begin{bmatrix}
        \bold{\Sigma}^{pos} &0\\
        0 & (\sigma^2)^{dir}
        \end{bmatrix}
        \bigg),    
\end{equation}
\begin{equation} \label{eq:damm_state}
        \Hat{\xi} = \begin{bmatrix}
        \xi^{pos}\\
        || \log_{\mu^{dir}}(\xi^{dir}) ||_2
    \end{bmatrix}\\
\end{equation}
where the state $\Hat{\xi}\in\mathbb{R}^{d+1}$ is augmented with the L2 norm of the Logarithmic mapping of the direction, which is not a unique value and varies relative to the directional mean of each component, the mean $\Hat{\mu}$ is padded with a $0$ as the Logarithmic mapping of $\mu^{dir}$ is always $0$ with respect to itself. The variance $(\sigma^2)^{dir}$ is appended along the diagonal in the new covariance $\Hat{\bold{\Sigma}}\in \mathbb{R}_{++}^{d+1}$ where all the off-diagonal entries are $0$ except the ones in $\bold{\Sigma}^{pos}\in \mathbb{R}_{++}^{d}$.

\begin{figure}[!t]
\centering
\includegraphics[width=1\linewidth]{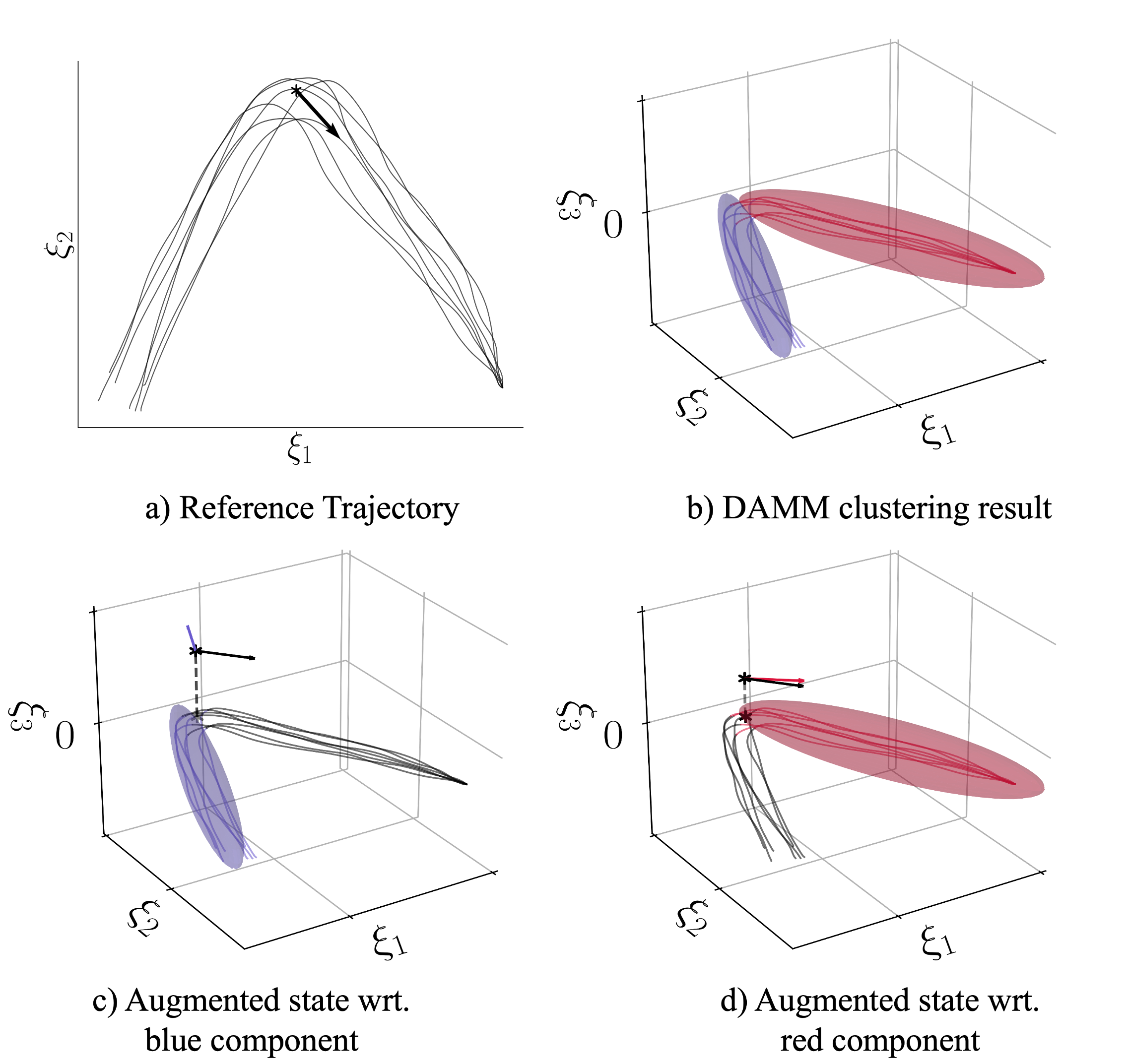}%
\hfill
\vspace{-5pt}
\caption{Illustration of DAMM: a) A-shaped reference trajectory and the point of interest marked in asterisk; b) clustering result of DAMM showing both clusters' covariance in ellipsoid; c) and d) overlay the point's direction in black and the directional mean of each component in color.}
\label{fig:damm}
\vspace{-10pt}
\end{figure}

We illustrate DAMM in the A-shaped trajectory from the 2D handwriting LASA dataset in Fig.~\ref{fig:damm}. Given the reference trajectory and a point of interest in a), we assign the point between the two components as shown in b). Although the original data is in 2D space, DAMM places each component at $\xi_3 =0$, and the 3D ellipsoids representing the covariance include $(\sigma^2)^{dir}$ in $\xi_3$ axis. When computing the probability of observing the point as in Eq.~\ref{eq:DAMM}, we augment the state relative to each component by Eq.~\ref{eq:damm_state}. Note that the larger deviation between our direction and the directional mean (blue) results in higher value in the additional dimension. On the contrary, when the direction is similar to the directional mean (red), the value in $\xi_3$ axis is closer to zero.

\subsection{DAMM Generative Model}

\label{sec:gen_damm}
Given the new Gaussian component in Eq.~\ref{eq:DAMM}, we now define the generative process of DAMM as follows:
\begin{equation}\label{eq:damm_model}
    \begin{gathered}
        \pi \sim GEM(1, \alpha), \\
        (\Hat{\mu}_k, \Hat{\Sigma}_k) \sim NIW(\Psi, \nu, \mu_0, \kappa),\\
        z_i \sim Cat(\pi_1, \pi_2, \dots),\\
        \Hat{\xi}_i | z_i=k \sim \mathcal{N}(\Hat{\mu}_k, \Hat{\Sigma}_k).
    \end{gathered}    
\end{equation}
Due to the variety and complexity of trajectory, it's intuitive for DAMM to automatically infer the number of components from the observed data. Hence, rather than drawing from a predefined fixed-length distribution, DAMM samples infinite-length cluster proportion, $\pi$ from the GEM (Griffiths Engen McCloskey) distribution following the stick-breaking process via the concentration factor $\alpha$~\cite{STICK-BREAKING}. DAMM then samples augmented Gaussian component from the conjugate prior NIW distribution, for which $(\Psi, \nu, \mu_0, \kappa)$ are the prior hyperparameters before seeing the data\cite{CONJUGATE}. We can then sample assignments $z_i$ from the categorical distribution defined by $\pi$, and observations of the augmented state $\Hat{\xi}_i$ from the newly defined Gaussian distribution as in Eq.~\ref{eq:DAMM}.

Using Bayesian conjugate prior allows us to incorporate prior belief in distribution. In particular, $\Psi\in \mathbb{S}_{++}^{d+1}$ (scale matrix) and $\nu\in \mathbb{R}_+$ (degrees of freedom) controls the variability of the covariance matrix $\bold{\Hat{\bold{\Sigma}}}$. In other words, we can regulate $(\sigma^2)^{dir}$ in Eq.~\ref{eq:sigma_dir} by tuning the hyperparameters. For example, given a nonlinear trajectory, if the prior belief is that $(\sigma^2)^{dir}$ is high, then a larger variation in direction is tolerated and DAMM will partition the trajectory into fewer linear components. And vice versa, meaning more components will be produced to respect the prior belief about a small variance.

\subsection{Parallel Sampling}
\label{sec:sampling}

Given the infinite-length cluster proportions in the DAMM generative model, using IW Gibbs sampler to infer and estimate the unknown parameters of DAMM could result in non-ergodic Markov chain; i.e., not every state can be visited and there are no guarantees of convergence because IW Gibbs sampler only instantiates a finite-length cluster proportion and cannot create new components as discussed in Section.~\ref{sec:gibbs}. However, the work in~\cite{PARALLEL} has proven that if an IW Gibbs sampler is mixed with any split mechanism that produces new components, then the resulting chain is indeed ergodic and the mixed sampler is a valid MCMC method. Hence, we introduce the efficient parallel sampling scheme that combines IW Gibbs sampler and Split/Merge Proposal together for the inference of the DAMM model. 

\textit{Split/Merge Proposal} was first introduced as an alternative MCMC method to Gibbs sampling for escaping low-probability local modes by moving groups of data points at once~\cite{SPLIT}. The original formulation of the Split/Merge Proposal, however, still employs CW Gibbs sampler to produce appropriate proposals, requiring incremental update and hindering parallel computation. We thus introduce a modified Split/Merge Proposal tailored to the IW Gibbs sampler.

In the context of the DAMM, we treat the assignment of each observation as latent variables and employ MCMC methods to draw a sample from the \textit{a posteriori} distribution. Hence, we designate $\mathbf{z} \in \mathbb{R}^N $ as the hidden state of the model which is a vector containing the assignments of $N$ augmented states $\hat{\xi}\in\mathbb{R}^{d+1}$. Say we are at a particular state $\mathbf{z}$ in the Markov chain, we can propose a candidate state $\mathbf{z}^*$ by performing either a split of one group or a merge between two groups, and then decide if the candidate proposal is accepted or not by evaluating the Metropolis-Hasting acceptance probability~\cite{METROPOLIS, HASTINGS},\begin{align} \label{eq:metropolis}
    a(\mathbf{z}^*,\mathbf{z}) = 
    \min \left[ 1,\ 
    \dfrac{q(\mathbf{z} \vert \mathbf{z^*})}{q(\mathbf{z}^* \vert \mathbf{z})} 
    \dfrac{\pi(\mathbf{z}^*)}{\pi(\mathbf{z})} \right],
\end{align}
\noindent where the target distribution $\pi(\mathbf{z})$ is the \textit{a posteriori} distribution $p(\mathbf{z} \vert \hat{\xi})$ we sample from, and the proposal distribution $q(\mathbf{z}^* \vert \mathbf{z})$ is the probability of reaching the candidate state $\mathbf{z}^*$ from the current state $\mathbf{z}$, i.e., $p(\mathbf{z}^* \vert \mathbf{z})$. The hat symbols are omitted hereinafter for the clarity of notation. 

We now look at each term in the context of \textbf{split} (See Appendix for \textbf{merge} proposal). As advised in~\cite{SPLIT}, random split or merge is highly unlikely to be accepted. Hence, we define a launch state $\mathbf{z}^{l}$ as the \textit{pseudo} current state in place of the original $\mathbf{z}$ in Eq.~\ref{eq:metropolis}. After choosing a component to split, we reach the launch state by randomly assigning the observations from the candidate component into two new components, and re-arranging the assignments via multiple scans of IW Gibbs sampler \textit{only} within the new components. From the launch state, we perform one \textit{final} scan of IW Gibbs sampler to reach the candidate state $\mathbf{z}^s$. Note that all other observations remain unchanged except the ones from the proposed component, that are split and re-arranged. If conjugacy is satisfied~\cite{CONJUGATE}, the ratio of the target distribution in Eq.~\ref{eq:metropolis} has the following analytical form:
\begin{equation}\label{eq:post}
    \dfrac{\pi( \mathbf{z}^{s})}{\pi(\mathbf{z})} 
    =
    \dfrac{
    \displaystyle{\prod_{z_i = z^{s}_1}} \pi_1 \mathcal{N}(\xi_{z_i} \vert \theta_{z^{s}_1})
    \prod_{z_i = z^{s}_2} \pi_2 \mathcal{N}(\xi_{z_i} \vert \theta_{z^{s}_2})}
    {\displaystyle{\prod_{z_i = z^{s}_{12}}}\mathcal{N}(\xi_{z_i} \vert \theta_{z^{s}_{12}})}
\end{equation}
where $z_{12}^s$ is the assignment of the proposed component, $z_{1}^s$ and $z_{2}^s$ are the assignments of the new groups \textit{after} the final scan of IW Gibbs sampler, $(\pi_1, \pi_2)$ are the cluster proportions, and $\mathcal{N}(\cdot \vert \theta_{z})$ is the \textit{a posteriori} probability distribution associated with assignment $z$ as in Eq.~\ref{eq:DAMM}.

The ratio of proposal distribution in Eq.~\ref{eq:metropolis} describes the probability of reaching the candidate state from the launch state by the \textit{final} scan of IW Gibbs sampler, yielding: 
\begin{equation} \label{eq:target}
     \dfrac{q(\mathbf{z} \vert \mathbf{z}^{s})}{q(\mathbf{z}^{s} \vert \mathbf{z})} = \prod_{z_i=z^{l}_{1}} \prod_{z_i=z^{l}_{2}}
     \dfrac
     {\pi_{z^{l}_1} \mathcal{N}(\xi_{z_i} \vert \theta_{z^{l}_{1}}) + \pi_{z^{l}_2} \mathcal{N}(\xi_{z_i} \vert \theta_{z^{l}_{2}})}
     {\pi_{z_i} \mathcal{N}(\xi_{z_i} \vert \theta_{z_i})   }
\end{equation}
where $z_{1}^l$ and $z_{2}^l$ are the assignments of the respective new groups \textit{before} the final scan of IW Gibbs sampler. Note that $q(\mathbf{z} \vert \mathbf{z}^{s})$ always has a probability of 1 because there is only one way of merging the two split groups into the original group. And $q(\mathbf{z}^{s} \vert \mathbf{z})$ is the product of the conditional probabilities of assigning the observations between the two new groups as in Eq.~\ref{eq:iw_gibbs}. 

\begin{figure}[!t]
\centering
\includegraphics[width=1\linewidth]{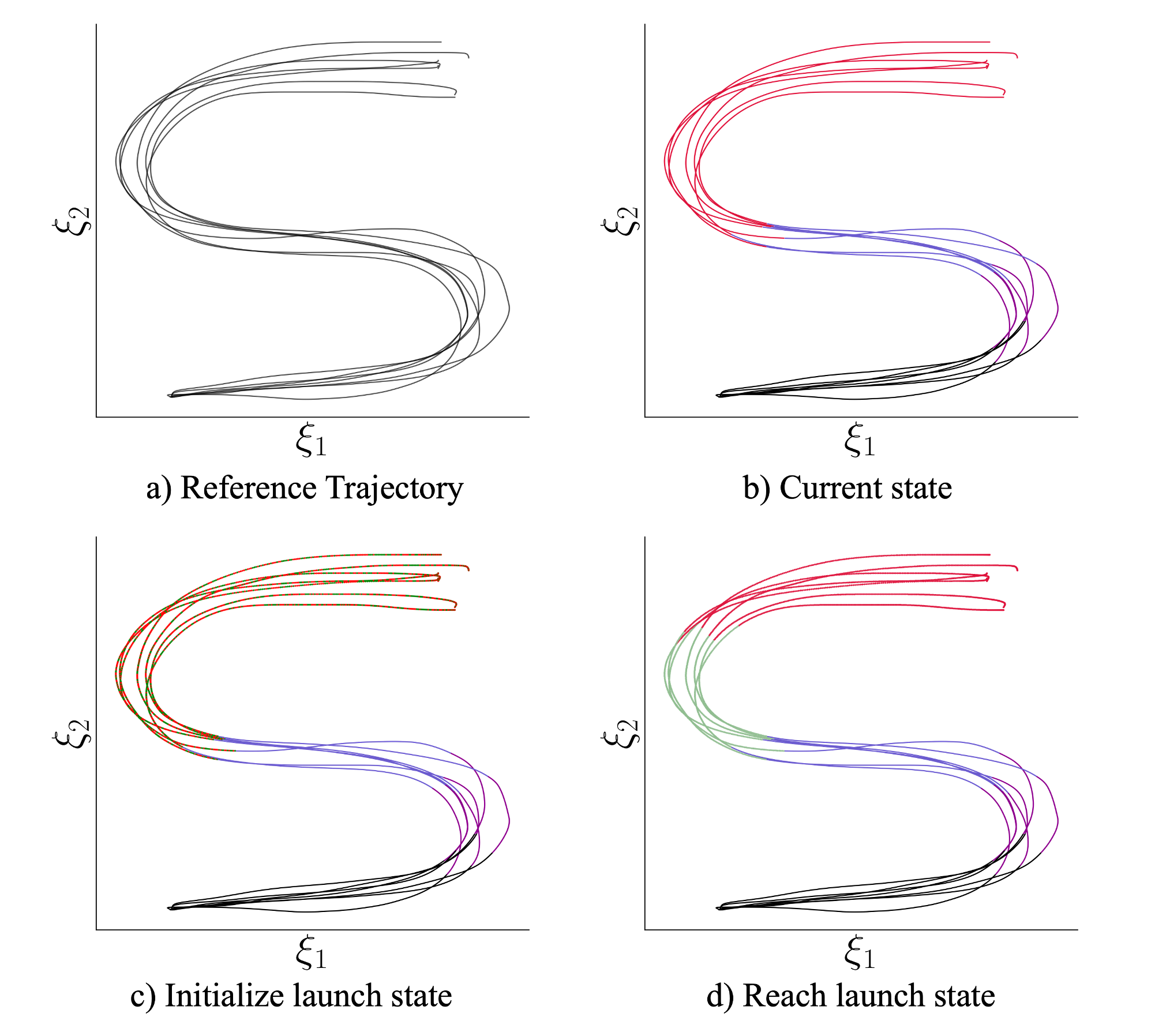}%
\vspace{-5pt}
\caption{Illustration of a split operation: a) an S-shaped reference trajectory; b) the current state of assignment; c) initialize the launch state by randomly assigning the candidate group (red) into two new groups (red and green); d) reach the launch state after multiple scans of IW Gibbs sampling.}
\label{fig:parallel_sampling}
\end{figure}

We illustrate an example of split operation in Fig.~\ref{fig:parallel_sampling}. Given a finite number of components, say $K=4$, IW Gibbs sampler reaches the current state of assignments in b). When a split is proposed for the red component, the launch state is initialized by randomly assigning the original group (red) into two new groups (red and green) in c). Multiple scans of IW Gibbs sampler are then performed only within the new groups, and the launch state is reached in d). We then perform one last scan of Gibbs sampling and evaluate the acceptance ratio to decide if the proposal is accepted or not.
\subsection{Mixed Sampler}
We have shown that the Split/Merge proposal is capable of producing new components, making it a well-suited complement to the IW Gibbs sampler for constructing an ergodic Markov chain. The combined sampler effectively alternates between the two MCMC methods as shown in Alg.~\ref{alg:alg1}. 

\setlength{\textfloatsep}{2pt}
\begin{algorithm}[!t]
\caption{Instantiated-Weight Parallel Sampling}\label{alg:alg1}
\begin{algorithmic}
\STATE  Initialize T \{Number of iterations\}
\STATE  Initialize $t \gets 0$
\STATE {\textbf{for }}$t = 0, \dots, T$ \textbf{do}
\STATE \hspace{0.5cm} Select a proposal randomly from $\{\textbf{Split}, \textbf{Merge}\}$
\STATE \hspace{0.5cm} Compute the launch state $\mathbf{z}^{l}_t$ by multiple scans of  
\STATE \hspace{0.5cm} IW Gibbs sampler by Eq.~\ref{eq:iw_gibbs}
\STATE \hspace{0.5cm} Reach the candidate state $\mathbf{z}^{*}_t$ by one \textit{final} scan 
\STATE \hspace{0.5cm} Evaluate the acceptance probability $a$ in Eq.~\ref{eq:metropolis}
\STATE \hspace{0.5cm} Select $a \sim U(0, 1)$
\STATE \hspace{0.5cm} \textbf{if} $a > \alpha$ \textbf{then}
\STATE \hspace{1 cm} $\mathbf{z}_t \gets \mathbf{z}_t^*\ \{\text{Accept proposal}\}$ 
\STATE \hspace{0.5cm} \textbf{else}
\STATE \hspace{1 cm} $\mathbf{z}_t \gets \mathbf{z}_t\ \{\text{Reject proposal}\}$ 
\STATE \hspace{0.5cm} $\mathbf{z}_{t+1} \gets$ IW Gibbs sampler by Eq.~\ref{eq:iw_gibbs}
\STATE $\textbf{return}\ \mathbf{z}$
\end{algorithmic}
\label{alg1}
\end{algorithm}

\vspace{-5pt}
\section{Experimental Results}
\label{sec:result}
\vspace{-5pt}
\subsection{Implementation}
\textbf{LPV-DS Estimation:} Recall that the LPV-DS parameters include the set of GMM parameters $\Theta_{\gamma}=\{\pi_k, \mu_k, \Sigma_k\}_{k=1}^K$ and the DS parameters $\Theta_{DS} = \{A_k, b_k\}_{k=1}^K$. DAMM estimates $\Theta_{\gamma}$ while the remaining DS parameters $\Theta_{DS}$ are estimated by solving the original semi-definite optimization problem introduced in \cite{PC-GMM} which minimizes the Mean Square Error (MSE) against the reference trajectories $\mathcal{D}$; i.e., $\min_{\theta_{DS}} J\left(\theta_{DS}\right)=\sum_{i=1}^{N}\left\|\dot{{\xi}}_{i}^{\mathrm{ref}}-{f}\left({\xi}_{i}^{\mathrm{ref}}\right)\right\|^2_2$ subject to the stability constraints defined in Eq. \ref{eq:lpv_ds}.

\textbf{Code:} DAMM is implemented in C++ with Python bindings and is available online with an efficient LPV-DS estimation at {\small \url{https://github.com/SunannnSun/damm}}
\subsection{Evaluation and Comparison}
\textbf{Datasets:} We conduct a comprehensive benchmark evaluation of the DAMM-based LPV-DS framework on the LASA handwriting dataset~\cite{SEDS} and the PC-GMM benchmark dataset~\cite{PC-GMM}. The LASA handwriting dataset contains a library of 30 human handwriting motions in 2D with single target, each containing 7 trajectories and totaling 7000 observations. The PC-GMM benchmark dataset consists of 15 motions characterized by highly non-linear patterns, featuring more complex behaviors than the LASA dataset. It includes 10 motions in 2D and 5 motions in 3D, with observations ranging from 800 to 3000 for each motion.

\textbf{Baselines:} We compare our approach against three different GMM estimation baselines: i) vanilla GMM on position (GMM-P), ii) vanilla GMM on position and velocity (GMM-PV), and iii) PC-GMM. \textit{Vanilla GMM} is referred to GMM inferred through standard Gibbs sampling.

\textbf{Evaluation metrics:} We perform an evaluation of our approach based on two categories: computational efficiency and model accuracy. We evaluate the computational efficiency by measuring the time each model takes to complete the inference given varying data size. The metrics on model accuracy are: \\
\begin{inparaenum}[(i)]
\item prediction root mean squared error $(\text{RMSE}) $:
\begin{equation}
            \text{RMSE} =  \frac{1}{N} \sum_{i=1}^N || \Dot{\xi}^{ref}_i-  f(\xi^{ref}_i)||,
\end{equation}
\item prediction cosine similarity or $\Dot{e}$:
\begin{equation}
            \Dot{e} = \frac{1}{N} \sum_{i=1}^N \left| 1 - \frac{f(\xi^{ref}_i)^T  \Dot{\xi}^{ref}_i}{||f(\xi^{ref}_i)|| ||\Dot{\xi}^{ref}_i||} \right|, 
\end{equation}
\item dynamic time warping distance as in~\cite{DTWD}:
\end{inparaenum}
\begin{equation}
    \textrm{DTWD} = \sum_{(i,j)\in\pi}d(\xi_i,\xi^{\textrm{ref}}_j),
\end{equation}
where $\pi$ is the alignment path between two time series, $i$ and $j$ are the sequence orders, and $d(\cdot,\cdot)$ measures the Euclidean distance between a pair of the series~\cite{DTWD}. RMSE and $\Dot{e}$ provide an overall assessment of the similarity between the resulting DS and the demonstration, and DTWD measures the dissimilarity between the reference trajectory and the corresponding reproduction from the same starting points.

\begin{figure}[!t]
    \centering
    \includegraphics[width=1\linewidth]{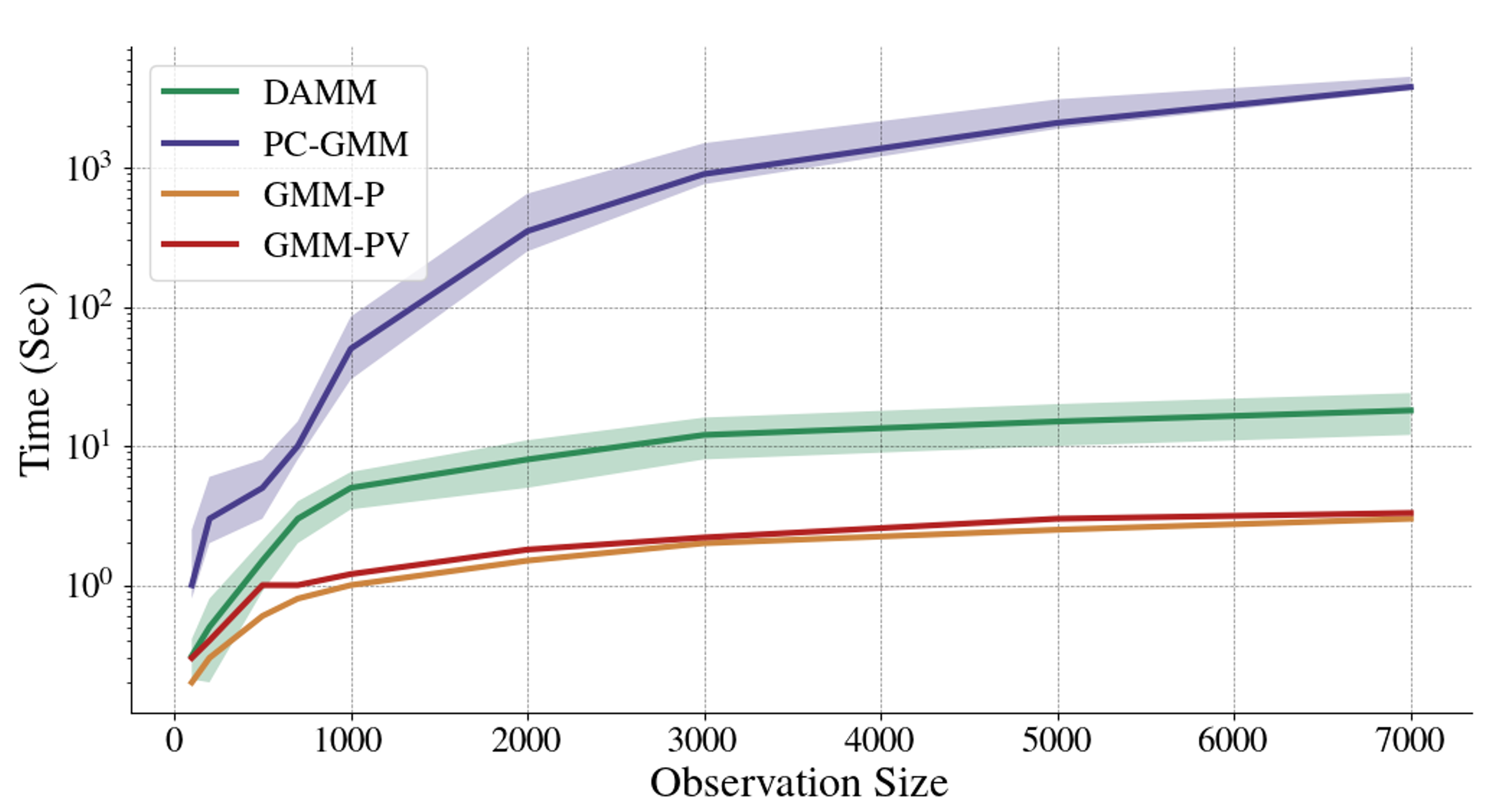}
    \caption{Comparison of computation time w.r.t varying observation size. All experiments are run on Ubuntu 20.04 with Intel i7-1065G7 @ 1.30GHz CPU and 16GB of RAM.} 
    \label{fig:time}
\end{figure}

\textbf{Results:} In Fig.~\ref{fig:time}, we measure the time each model takes to complete the training and inference across varying data size. We note that when learning a small-sized trajectory ($< 500$ observations), the distinctions in computation time are non-significant as all four methods can finish within 10 seconds. The distinction, however, becomes more pronounced when dealing with larger datasets. For example, given an average demonstration containing 7000 observations from LASA dataset, DAMM scales well with large datasets and completes the clustering task slightly over 10 seconds. On the contrary, the computation time of PC-GMM grows exponentially and it takes more than an hour to complete the task due to its \textit{strictly sequential} nature. We note that DAMM falls behind the vanilla-GMM in speed mostly due to the iterative computation of directional mean as in Eq.~\ref{eq:riem_mu_sigma} and the intermediate Gibbs sampling scans required to reach the launch state as discussed in Section.~\ref{sec:sampling}. Nevertheless, DAMM still achieves significant speedup in computational speed by order of magnitude compared to PC-GMM (our goal). 


\begin{table}[!t]
\caption{Comparison of the average performance between DAMM and baselines over the entire LASA dataset and PC-GMM dataset.\label{tab:table1}}
\scriptsize
\centering
\begin{tabular}{cc|ccc}
\hline \hline
 & \multirow{2}{*}{Model} & &\textbf{Model Accuracy} & \\
 & & RMSE & $\Dot{e}$ & DTWD\\ \hline
\multirow{3}{1.3cm}{\centering \textbf{PC-GMM Dataset}} & GMM-P & $1.2\pm 0.6$ & $0.35\pm 0.19$ & $569\pm 89$\\
 &GMM-PV &  $1.5\pm 1.3$ & $0.52 \pm 0.22$ & $692\pm 94$ \\
 &PC-GMM & $1\pm 0.4$ & $\mathbf{0.07\pm 0.03}$ & $313\pm 28$\\
 & \cellcolor{LightCyan} DAMM & \cellcolor{LightCyan} 
 $\mathbf{0.9\pm0.3}$ & \cellcolor{LightCyan} 
 $\mathbf{0.07\pm 0.03}$ & \cellcolor{LightCyan} 
 $\mathbf{295\pm 20}$ \\
 \hline   \hline
\multirow{4}{1.3cm}{\centering \textbf{LASA Dataset}} & GMM-P & $1.28\pm 0.68$ & $0.36\pm 0.20$ & $581\pm 99$\\
 & GMM-PV & $1.38\pm 1.02$ & $0.48 \pm 0.20$ & $690\pm 91$ \\
 & PC-GMM & $0.96\pm 0.39$ & $0.09 \pm 0.04$ & $331\pm 39$\\
 & \cellcolor{LightCyan} DAMM & \cellcolor{LightCyan} $\mathbf{0.81\pm0.23}$ & \cellcolor{LightCyan} $\mathbf{0.07\pm0.02}$ & \cellcolor{LightCyan} $\mathbf{280\pm 20}$\\ \hline \hline
\multicolumn{4}{l}{$^*$The optimal results are marked in bold.}
\end{tabular}
\end{table}

\begin{figure*}[!t]
\centering
\includegraphics[width=0.95\linewidth]{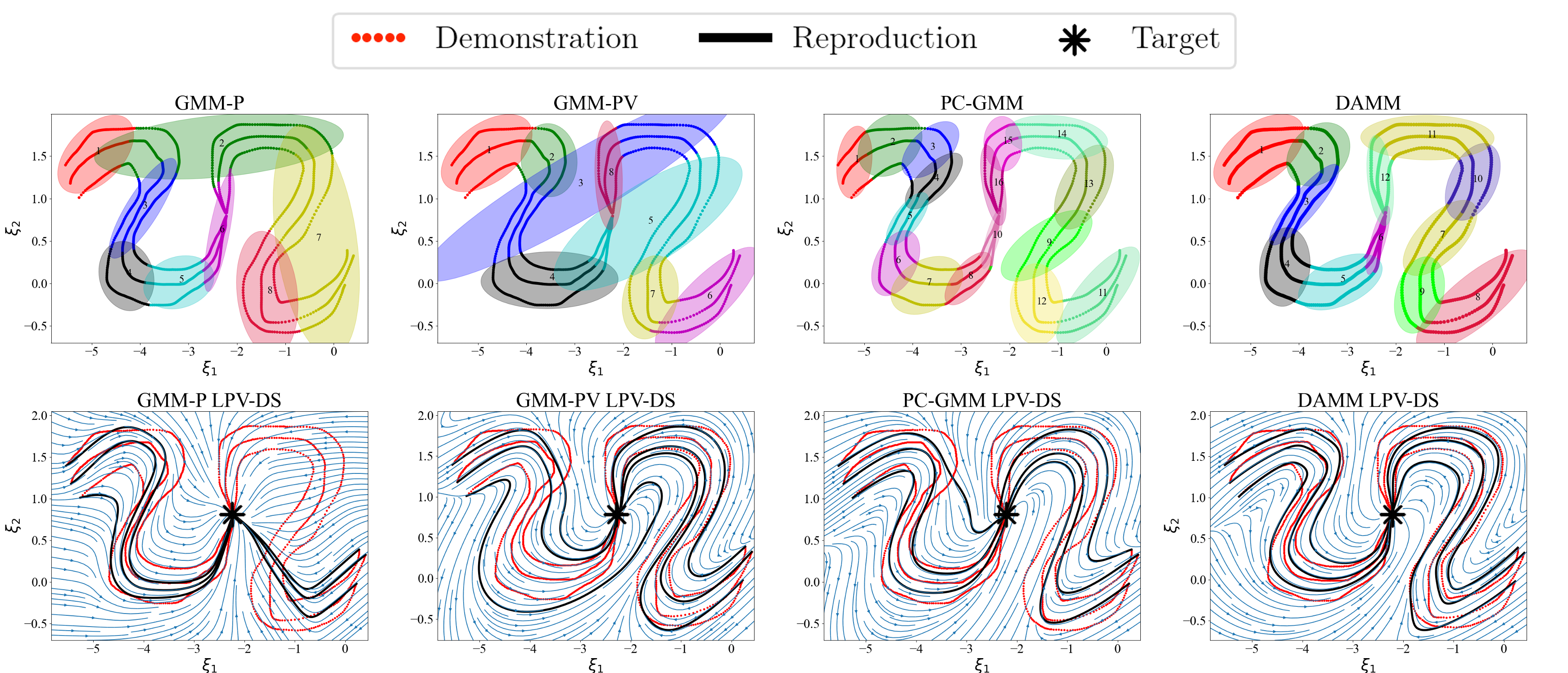}%
\caption{Comparison of clustering (top) and reproduction (bottom) results between the GMM-P (position only) LPV-DS, GMM-PV (position+velocity) LPV-DS, PC-GMM LPV-DS and \textbf{DAMM LPV-DS} on a multi-behavior trajectory obtained from~\cite{PC-GMM}. Both GMM-P and GMM-PV are fitted via Gibbs Sampling. The \textbf{computation times are 0.3, 0.5, 53 and 1.2 in seconds from left to right}. Notice the improved reproduction accuracy resulting from optimal GMM fitting via DAMM on the right column.}\label{fig_sim}
\label{fig:multi}
\end{figure*}

Table.~\ref{tab:table1} compares the performance between DAMM and baseline methods across the three metrics in the category of model accuracy. We note that DAMM outperforms the vanilla GMM methods across all three metrics (lower the better) in both benchmark datasets, and holds a slight edge over PC-GMM in RMSE and DTWD. The non-significant difference between DAMM and PC-GMM in performance is expected, as both methods effectively capture the directionality and generate appropriate models, leading to proper DS via optimization. A comparison result between four methods on a 2D \textit{multi-behaviour} trajectory from PC-GMM dataset is shown in Fig.~\ref{fig:multi}. Note both GMM-P and GMM-PV fail to capture the directionality of the motion, consequently producing erroneous DS. On the contrary, both DAMM and PC-GMM approach the non-linearity of the trajectory by identifying and clustering the linear portions, producing physically-meaningful representation of the trajectory.

\begin{figure}[!t]
    \centering
    \includegraphics[width=0.95\linewidth]{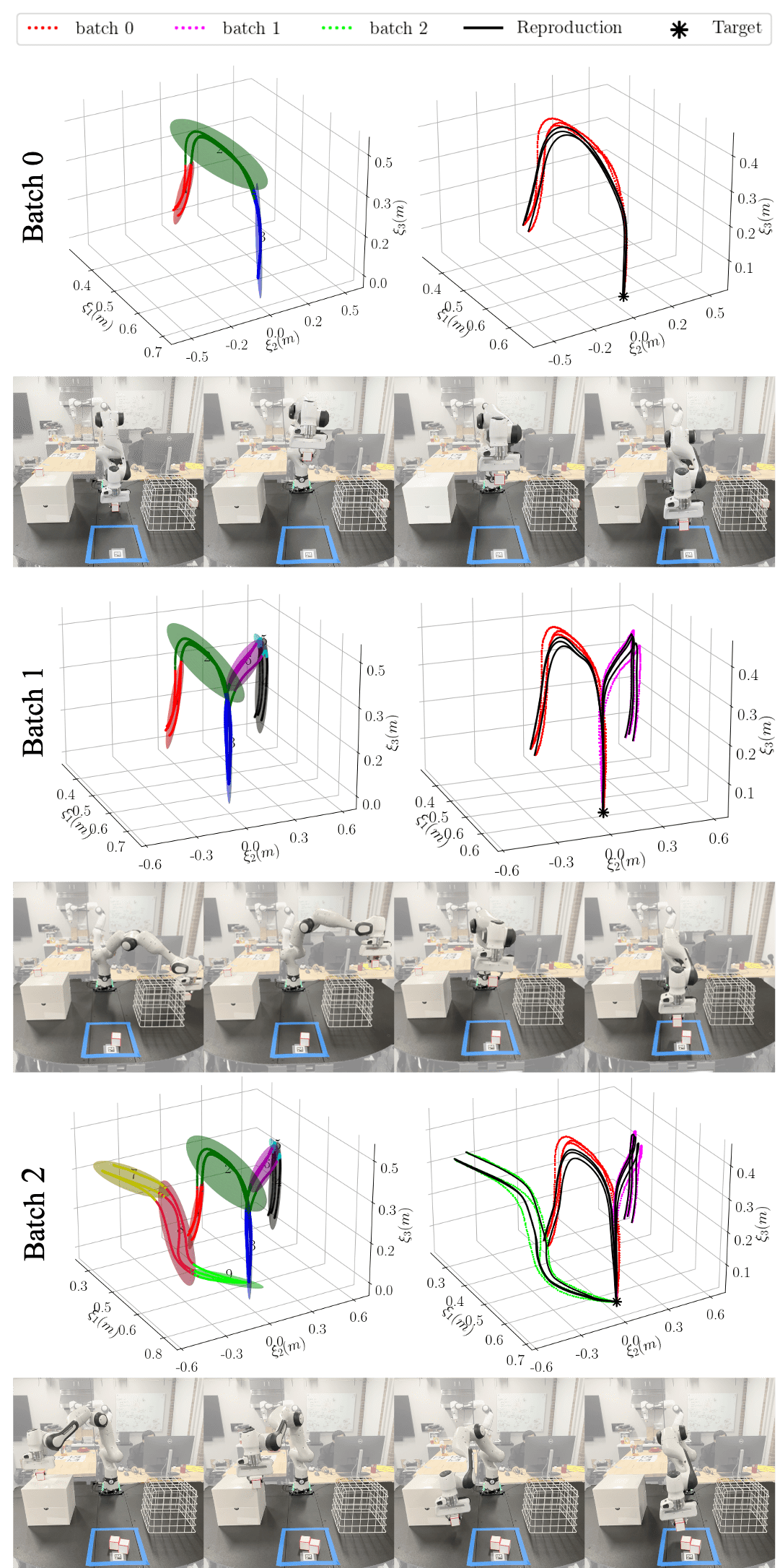}
    \caption{Sequence of learning three tasks incrementally via the DAMM-based LPV-DS. Every two rows correspond the learning of a new task with the clustering result (top left), the reproduction DS (top right), and the snapshot sequence of the execution (bottom)\label{fig:increm}. The \textbf{computation times are 0.6, 0.5 and 0.5 in seconds respectively for each task}.}
\end{figure}

\subsection{Robot Experiments}\label{sec:robot}
We validate our approach on a Frank Emika Panda robot in the application of incremental learning where the trajectory data, provided kinesthetically by humans, comes in progressively. The traditional approach is to concatenate batches and re-learn the combined trajectory, resulting in inefficient use of data. We therefore propose an alternative approach where new data can choose to either join existing components or form new groups while keeping the assignment of the previous batches unchanged. This efficiently reduces the task to clustering only the new data, circumventing the need to learn the combined batch.

 
In Fig.~\ref{fig:increm}, we showcase the compatibility of DAMM in our new incremental learning framework where the robot sequentially learns three different tasks. Each task comprises 3 demonstrations and approximately 500 observations. Upon receiving the demonstration batch 0, the robot performs DAMM-based LPV-DS with the clustering and the reproduction results shown in the first two rows of Fig.~\ref{fig:increm}. Subsequently, we introduce another demonstration batch 1, which moves the object from different locations but later merges with batch 0. The middle two rows illustrate that the new demonstration initially forms distinct components but later joins the first demonstration as both batches converge. The reproduction DS and the snapshot sequence confirms the robot's successful learning and execution of the new trajectory while preserving the preceding DS. When the last demonstration batch 2 comes in, with no overlapping with the previous batches, the last two rows show that batch 2 forms its own groups and the robot successfully executes the newly learned DS by moving the object to the target location via a different trajectory. We highlight that DAMM learns each batch in less than 1 second, and produces physically-meaningful clustering results with reliable DS for the robot to reproduce the demonstration trajectory and reach the target at near real-time scale.

\section{Conclusion}
\label{sec:conclusion}
We introduce the Directionality-Aware Mixture Model that is capable of effectively identifying the directional features of the trajectory data. By including both the positional and directional information using a proper Riemannian metric, DAMM produces physically-meaningful clustering results that represent the intrinsic structure of the trajectory data. Along with the parallel sampling scheme, the DAMM-based LPV-DS framework achieves a drastic improvement in computational efficiency while remaining comparable to the state-of-the-art level of model accuracy. However, we note that DAMM was formulated on positional data only. For more adaptive motion policy, an integration of DAMM with orientation control is necessary.


\appendix[Merge Proposal]
We define the launch state $\mathbf{z}^l$ by randomly initializing the assignments between two groups of interest and performing multiple scans of IW Gibbs sampler. We then compute the expressions below as if we are reaching the original split state from the $\mathbf{z}^l$ by one \textit{final} scan of Gibbs sampler:
\begin{equation}
    \dfrac{\pi( \mathbf{z}^{m})}{\pi(\mathbf{z})} 
    =
    \dfrac{\displaystyle{\prod_{z_i = z^{m}_{12}}}\mathcal{N}(\xi_{z_i} \vert \theta_{z^{m}_{12}})}
    {\displaystyle{\prod_{z_i = z^{m}_1}} \pi_1 \mathcal{N}(\xi_{z_i} \vert \theta_{z^{m}_1})
    \prod_{z_i = z^{m}_2} \pi_2 \mathcal{N}(\xi_{z_i} \vert \theta_{z^{m}_2})}
\end{equation}
\begin{equation}
     \dfrac{q(\mathbf{z} \vert \mathbf{z}^{m})}{q(\mathbf{z}^{m} \vert \mathbf{z})} = \prod_{z_i=z^{l}_{1}} \prod_{z_i=z^{l}_{2}}
     \dfrac
     {\pi_{z_i} \mathcal{N}(\xi_{z_i} \vert \theta_{z_i}) }
     {\pi_{z^{l}_1} \mathcal{N}(\xi_{z_i} \vert \theta_{z^{l}_{1}}) + \pi_{z^{l}_2} \mathcal{N}(\xi_{z_i} \vert \theta_{z^{l}_{2}})}
\end{equation}
where $z^{m}_1$ and $z^{m}_2$ are the respective assignment labels of the original components, and $z^{m}_{12}$ is the new assignment label of the combined component. $z^{l}_1$ and $z^{l}_2$ are the assignments of the two groups before the final scan of IW Gibbs sampler. To propose meaningful merge between appropriate groups, we pick two candidates using metrics such as Gaussian similarity and Euclidean distance between means.

\bibliographystyle{IEEEtran}
\bibliography{reference}

\end{document}